\documentclass[conference]{IEEEtran}
\IEEEoverridecommandlockouts
\usepackage{cite,soul}
\usepackage{amsmath,amssymb,amsfonts,mathrsfs}
\usepackage{algorithmic}
\usepackage{graphicx}
\usepackage{multirow,booktabs}
\usepackage{textcomp}
\usepackage{xcolor,xspace}
\usepackage{CJK}
\usepackage{textcomp,xspace}
\usepackage{hyperref}
\usepackage{marvosym}
\def\BibTeX{{\rm B\kern-.05em{\sc i\kern-.025em b}\kern-.08em
    T\kern-.1667em\lower.7ex\hbox{E}\kern-.125emX}}
\begin{document}

%define
\newcommand{\icaf}{\emph{I}CAF\xspace}

\title{\textbf{\icaf}: Iterative Contrastive Alignment Framework for Multimodal Abstractive Summarization}

\author{
    \IEEEauthorblockN{Zijian Zhang$^{\star,a}$\thanks{$\star$ These authors contributed equally to this work.}, Chang Shu$^{\star,b,c}$, Youxin Chen$^b$, Jing Xiao$^b$, Qian Zhang$^c$ and Lu Zheng$^{c, \href{mailto:zheng.lu@nottingham.edu.cn}{\textrm{\Letter}}}$} 
    \IEEEauthorblockA{$^a$ \textit{Meituan, Shanghai, China} \\
                      $^b$ \textit{Ping An Technology (Shenzhen) Co., Ltd, Shenzhen, China} \\
                      $^c$ \textit{University of Nottingham Ningbo China, Ningbo, China}}
    \IEEEauthorblockA{zijian.zh96@gmail.com, \{scxcs1, qian.zhang, zheng.lu\}@nottingham.edu.cn, \{chenyouxin149, xiaojing661\}@pingan.com.cn}
}

\begin{CJK}{UTF8}{gbsn}
\maketitle

\begin{abstract}
Integrating multimodal knowledge for abstractive summarization task is a work-in-progress research area, with present techniques inheriting fusion-then-generation paradigm. Due to semantic gaps between computer vision and natural language processing, current methods often treat multiple data points as separate objects and rely on attention mechanisms to search for connection in order to fuse together. In addition, missing awareness of cross-modal matching from many frameworks leads to performance reduction. To solve these two drawbacks, we propose an Iterative Contrastive Alignment Framework (\icaf) that uses recurrent alignment and contrast to capture the coherences between images and texts. Specifically, we design a recurrent alignment (RA) layer to gradually investigate fine-grained semantical relationships between image patches and text tokens. At each step during the encoding process, cross-modal contrastive losses are applied to directly optimize the embedding space. According to ROUGE, relevance scores, and human evaluation, our model outperforms the state-of-the-art baselines on MSMO dataset. Experiments on the applicability of our proposed framework and hyperparameters settings have been also conducted.
\end{abstract}


% section-1
\section{Introduction}
\begin{figure}[t]
  \centering
  \includegraphics[width=0.76\linewidth]{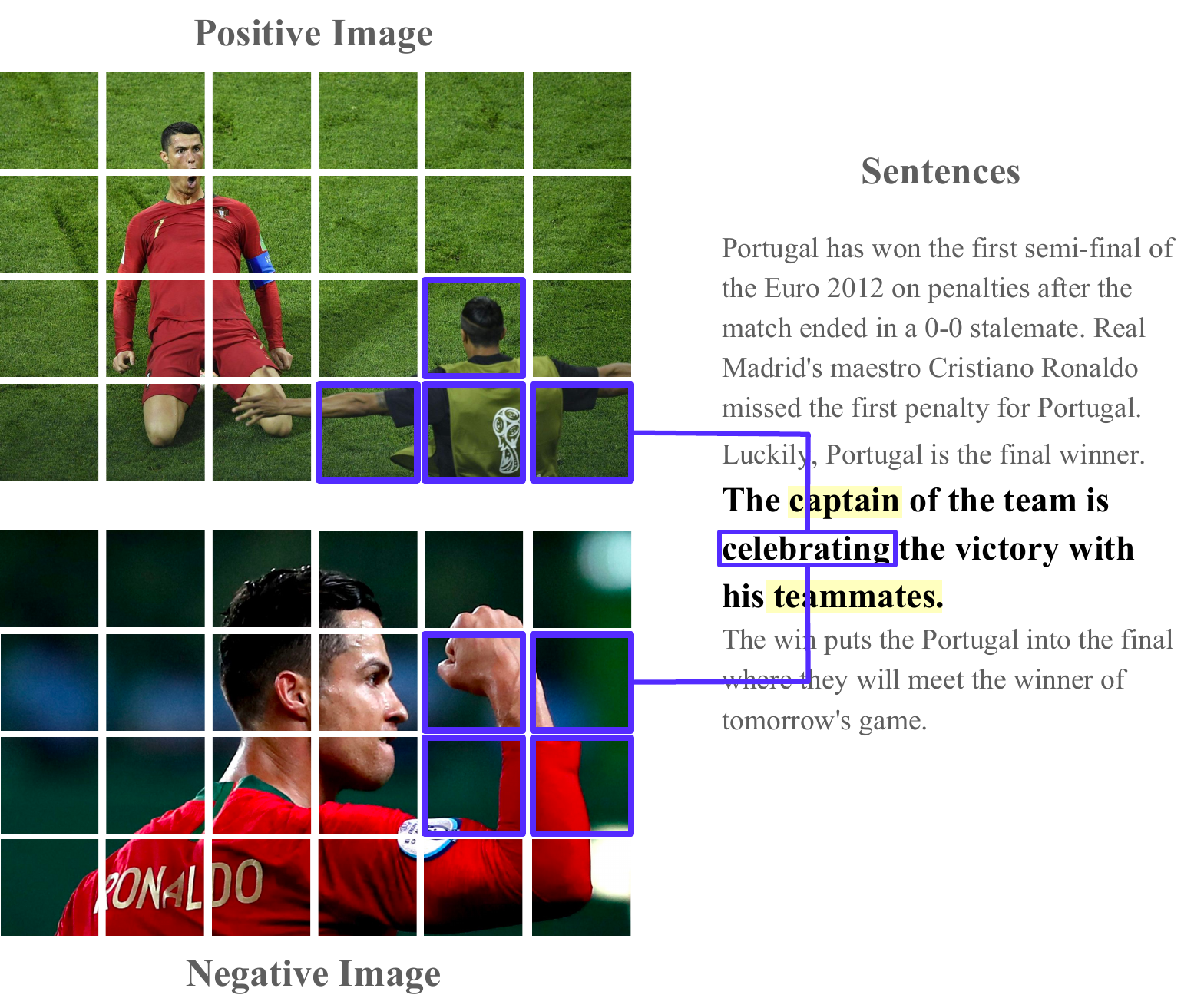}
  \caption{An example of cross-modal semantical alignment.}
  \label{fig:example}
\end{figure}

The goal of Multimodal Abstractive Summarization (MAS) is to compress information from interacting modalities into a brief, simple, and legible summary \cite{liu-etal-2020-multistage}. With the growing number of videosharing platforms and multimodal data, MAS can help consumers rapidly obtain the information they need. The difficulty of tackling MAS has been alleviated with recent advancement in multimodal fusion and text generation  \cite{liu-lapata-2019-text}, whose classical scheme \cite{libovicky-helcl-2017-attention} combining language and vision features to generate textual summary based on seq2seq hierarchical attention methods. Despite promising results have been obtained, we discovered that current techniques still remain the following drawbacks:

\textbf{Comprehension gap between vision and language.} As seen in Figure~\ref{fig:example}, it is natural for human to focus more on the word \emph{celebrating} that is caused by the semantic corresponding regions in the image (indicated by blue boxes). Since current approaches use different structures for multimodal data in vision and language \cite{lee2018stacked, li2020aspect}, i.e., the inconsistency between image and text models makes them less concerned with the deep semantics of each cross-modal corresponding counterpart segment. Large-scale text generation models based on transformers, such as UniLM \cite{dong2019unified}, BART \cite{lewis-etal-2020-bart}, or ProphetNet \cite{qi2020prophetnet}, have recently demonstrated outstanding performance on generation tasks \cite{tan2019lxmert}. Although leveraging and adapting these techniques to MAS is still an ongoing research topic that yet to be fully explored \cite{zhu2020multimodal}, there have been several attempts. For example, CtnR \cite{9534082} aimed to unsupervisedly align various modalities while merely capturing shallow information. Using selected encoders to represent modalities and summarize information, Li et. al. \cite{li-etal-2020-multimodal} attempted to bridge the gap but struggle to produce image-text corresponding captions. Zhu et. al. \cite{zhu2021graph} tackled MAS task with graph based intra modal reasoning method. Oscar \cite{li2020oscar} integrated the contrastive learning to learn cross-modal representation on image-text pairs, but it is contingent on the a priori condition that the object tags in image can be accurately detected and appear in the text. To summarize, existing attention-based or reasoning-based models have difficulty in capturing the fine-grained pairwise associations among large number of region-word fragment pairs because of the obvious heterogeneity gap between image and text features. In our opinion, using comparable model structures to unify the representation and shrink the semantic gaps between vision and language is an effective strategy.
% figure 1

\textbf{Lack of awareness of image-text matching.} Natural matches exist between each other due to the pairing of images and sentences in the existing dataset \cite{sanabria18how2, zhu-etal-2018-msmo}. While being beneficial to the training process, this decreases the generalization of the models and inhibits further model performance improvements\cite{sanabria2019deep}. If the words \emph{caption} and \emph{teammates} (yellow surrounding) are not taken into account, the sentences that correspond to two images in Figure~\ref{fig:example} may believe that the bottom image is a better one and compatible with the phrases. Li et. al. \cite{li2018read} take image-text matching as a priori step before multimodal summarization to solve this problem. In essence, this prevents the model from distinguishing useful data from those are irrelevant. Entity matching is also considered in the abstractive generation via bottom-up attention \cite{10.1145/3183713.3196926}, where entities should be well described otherwise performance would suffer. In a fine-grained task, i.e., cross-modal retrieval \cite{huang2019attention}, hard triplet loss \cite{faghri2018vse++} is used to enhance the corresponding representation across vision and language. With similar ideas, MultimodalSum \cite{mm2021} proposes to generate summary from multiple reviews in a self-supervised manner. In general, these methods have focused on the awareness of matching, but have not been attempted in multimodal summarization, especially the semantic matching from bottom up.

Motivated by the potential of solving these issues, this paper seeks to guide multimodal summarization with iterative alignment and progressively matching corresponding image-text pairs through contrastive learning, termed \icaf. Specifically, we first propose a recurrent alignment (\textbf{RA}) layer that investigates the correspondence between vision and language for summarization via two modules: 1) an iterative cross-modal attention module (\textbf{CAM}) to align fragments across image and text; 2) a renovation addition module (\textbf{RAM}) to integrate the current alignment with history knowledge. The iterative alignment approach may \emph{gradually} update cross-modal attention in order to amass clues for identifying matching semantics and improving cross-modal information interaction. However, employing only attention-based mechanism to align cross-modal semantics yields inadequate results. As a solution, contrastive losses \cite{oord2018representation} are applied after each RA layer to keep the embedding space constant and maintain distance from other references. Aggregating multiple contrastive objectives may also be used to directly supervise the learning of image-text correspondences, while accelerating model convergence and increasing performance at the same time. By taking aforementioned proposed strategies, we show that \icaf can produce a more illustrative summary and achieve the state-of-the-art results empirically. In conclusion, the contributions of this paper are:
\begin{itemize}[]
\setlength{\itemsep}{3pt}
\setlength{\parsep}{1pt}
\setlength{\parskip}{1pt}
\item a modified recurrent alignment layer in the encoder is proposed to handle the complexity of semantics, which incorporates CAM for aligning information and RAM for aggregating knowledge.
\item two auxiliary contrastive losses are used in encoder, with image-to-text and text-to-image objectives, respectively.
\item we thoroughly verify our \icaf on benchmark dataset and evaluate the settings of hyperparameters and novel units in details.
\end{itemize}

% section-2
\section{Methodology}
% figure 2
\begin{figure*}[t]
\centering
\includegraphics[width=\linewidth]{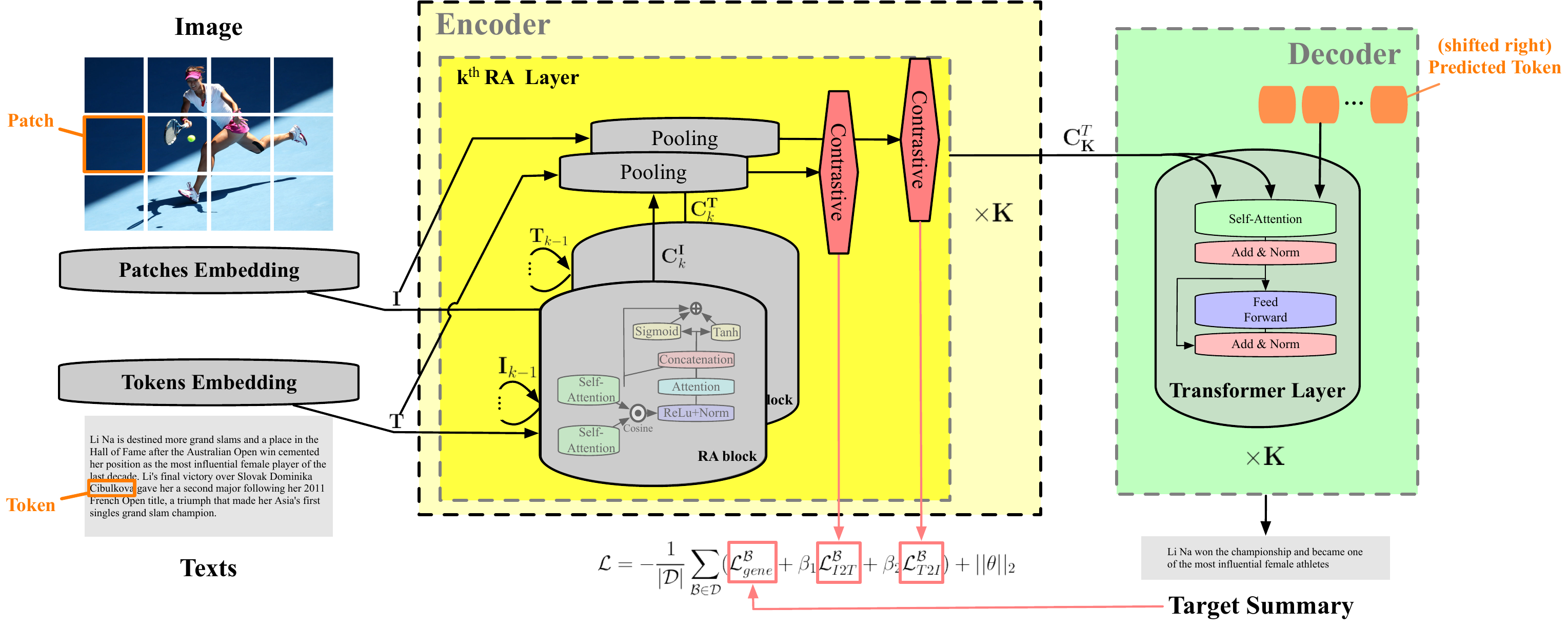}
\caption{\label{fig:frame} Framework of the proposed \icaf. The details of embedding are introduced in Section~\ref{sec:represent}. The whole structure consists of an encoder to align and fuse multimodal features and a standard decoder to reconstruct the summary sequence, which is trained by a combination of triple losses.}
\end{figure*}

In this section, we will elaborate on the details of our proposed \icaf for multimodal abstractive summarization task. First, we introduce our way of multimodal (Image\&Text) feature representation in Section~\ref{sec:represent}. Then, in Section~\ref{sec:ra}, we show the structure of RA and how it can be implemented into the \icaf as a module (or layer). The interactive contrastive based InfoNCE losses between text and image embeddings, are explained in Section~\ref{sec:frame}. Finally, the total objective function is described in Section~\ref{sec:objectives}.

\subsection{Image\&Text Representation}
\label{sec:represent}
\textbf{Text representation} To achieve the fine-grained connection between vision and language, we extract word-level features by combining textual tokens with positional embeddings. Specifically, for a given sentence with $N$ words, we firstly project the sentence with a contiguous learnable embedding matrix $\textbf{W}_{tok} \in \mathbb{R}^{V \times D}$, where $V$ and $D$ are vocabulary size and embedding dimension respectively. Together with statistic positional embedding \cite{vaswani2017attention}, the text representation is denoted by $\textbf{T} \in \mathbb{R}^{N \times D}$.

\textbf{Image representation}
Conventionally CNN-based image representation may not work well in the transformer module \cite{wu2021cvt, d2021convit}. Encouraged by \cite{dosovitskiy2021an}, we reshape the original image $\textbf{I} \in \mathbb{R}^{C\times H \times W}$ ($C,H,W$ denote number of channels, height, and width, respectively) into a sequence of flattened 2D patches $\textbf{I} \in \mathbb{R}^{M\times D'}$ using patches projection. $M$ denotes the number of regions (or patches) and $D'$ denotes embedding dimension for vision representation. In order to maintain the comparability of image and text features in the subsequent operations, we map the original dimension $D'$ to $D$ by learnable projection.

\subsection{Recurrent Alignment Layer}
\label{sec:ra}
% figure 3
\begin{figure}[t]
\centering
\includegraphics[width=\linewidth]{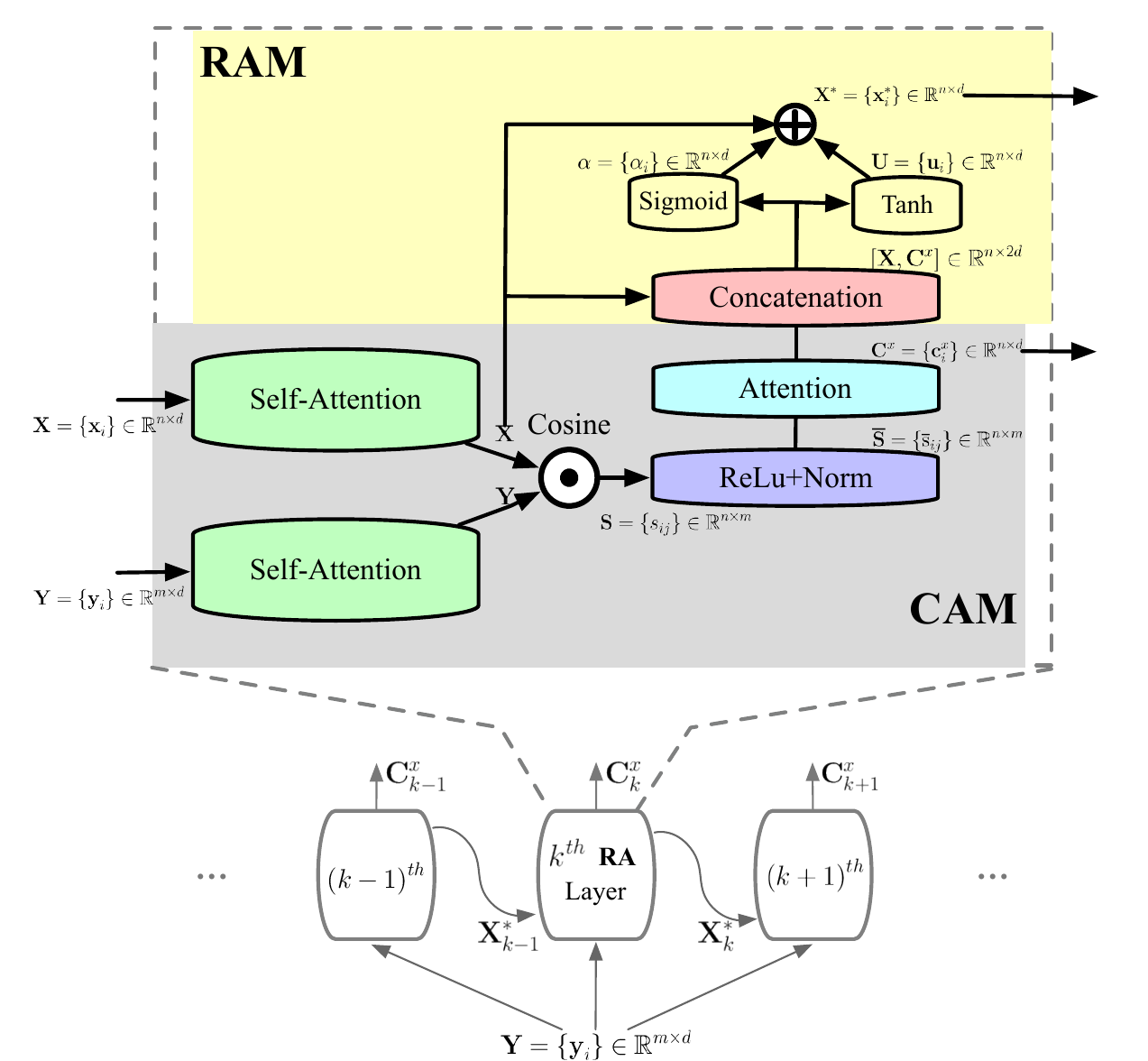}
\caption{\label{fig:ra} The visualization of recurrent flow (bottom) and structure of RA block (top), which are generalized for $\textbf{X}$ and $\textbf{Y}$.}
\end{figure}

The encoder of \icaf consists of multiple RA layers that aim to align fragmental features (token or patch embeddings) by refining the \emph{current} state and \emph{previous} knowledge in a recurrent manner. The former seizes the corresponding information between image and text features and the latter discards trivial knowledge. RA can be regarded as a block (see Figure~\ref{fig:frame}) that takes two inputs, \emph{i.e.,} $\textbf{T}$ and $\textbf{I}$. For current state, RA focuses on the alignment among input pair by CAM. For previous knowledge, RA concerns more about how to update the state by RAM.

For generalization, we denote the two input sets of features as $\textbf{X}=\{\textbf{x}_i | i=1,..,n\}$ and $\textbf{Y}=\{\textbf{y}_i | i=1,..,m\}$, where $\textbf{x}_i \in \mathbb{R}^{d}$ and $\textbf{y}_i \in \mathbb{R}^{d}$ have the same hidden dimension. Note that $\textbf{X}$ can be either $\textbf{T}$ or $\textbf{I}$, while $\textbf{Y}$ is the other. The visualization flows described below is also shown in Figure~\ref{fig:ra}.

\textbf{Cross-modal Attention Module (CAM).} The CAM aims to account for global information in $\textbf{Y}$ for each local feature $\textbf{x}_i$ in $\textbf{X}$. To achieve this goal, we first compute the cosine similarity between each pair $(\textbf{x}_i, \textbf{y}_j)$ followed by a standard self-attention module to enhance inter information:
\begin{equation}
    {\rm s}_{ij} =\frac{\textbf{x}^{T}_{i}\textbf{y}_{j}}{||\textbf{x}_i|| \cdot ||\textbf{y}_j||}
\end{equation} As in \cite{lee2018stacked}, we further normalize the similarity score $s$ and filter high relevance items by adding an adjustable factor $\gamma$:
\begin{equation}
    \overline{\rm s}_{ij} = \frac{{\rm relu}({\rm s}_{ij}+\gamma)}{\sqrt{\sum_{i=1}^n}({\rm s}_{ij}+\gamma)^2}
\end{equation} where ${\rm relu=max}(x,0)$.

To seize the alignment information from $\textbf{Y}$ to $\textbf{X}$, attention-based mechanism \cite{attn2015} is applied by:
\begin{equation}
    \begin{aligned}
        \textbf{c}^{x}_i & = \sum_{j=1}^{m} {\rm \alpha}_{ij}\textbf{y}_j \\
        s.t. \quad {\rm \alpha}_{ij} &= \frac{exp(\lambda \overline{\rm s}_{ij})}{\sum_{j=1}^{m} exp(\lambda \overline{\rm s}_{ij})}
    \end{aligned}
\label{fm:c}
\end{equation} where $\lambda$ is the inverse temperature of the softmax function \cite{10.5555/2969239.2969304} to adjust the smoothness of the attention distribution.

We define $\textbf{C}^x = \{\textbf{c}^{x}_i|i=1,...,n\} \in \mathbb{R}^{n \times d}$ as $\textbf{X}$-target alignment features. Each element $\textbf{c}^{x}_i$ captures related semantics contributed by the reference feature $\textbf{Y}$.

\textbf{Renovation Addition Module (RAM).} To refine the alignment feature for the next layer in the encoder, which updates the current state by previous knowledge, RAM is to aggregate the input feature $\textbf{X}$ with alignment feature $\textbf{C}^x$ dynamically:
\begin{equation}
    \textbf{x}^{*}_i = f(\textbf{x}_i, \textbf{c}^{x}_i)
\end{equation} where $f()$ is an aggregating function. We adopt a gate mechanism \cite{li-etal-2018-self} for $f()$:
\begin{equation}
    \begin{aligned}
    \alpha_i &= {\rm sigmoid}(\textbf{W}_{\alpha}[\textbf{x}_i,\textbf{c}^{x}_i]+\textbf{b}_{\alpha}) \\
    \textbf{u}_i &= {\rm tanh}(\textbf{W}_u[\textbf{x}_i,\textbf{c}^{x}_i]+\textbf{b}_u) \\
    \textbf{x}^{*}_i &= (1-\alpha_i) * \textbf{x}_i + \alpha_i * \textbf{u}_i
    \end{aligned}
\label{fm:x}
\end{equation} where $\textbf{W}_{\alpha}, \textbf{W}_{u}, \textbf{b}_{\alpha}, \textbf{b}_u$ are learnable parameters. $\textbf{u}_i$ is a fused feature, which enhances the interactive information between previous knowledge $\textbf{x}_i$ and the current state $\textbf{c}^{x}_i$. And $\alpha_i$ is a forgetting coefficient that is able to filter trivial information in previous knowledge $\textbf{x}_i$ and focus more on the alignment information shared from $\textbf{Y}$.

\textbf{Recurrent Alignment (RA) Layer.} We integrate CAM and RAM into a RA layer, whose k$^{th}$ layer outputs are:
\begin{equation}
    \textbf{C}^{x}_k, \textbf{X}^{*}_k = \textbf{RA}(\textbf{X}^{*}_{k-1},\textbf{Y}), k=1,...,K
\end{equation} where the details of $\textbf{C}^{x}_k$ and $\textbf{X}^{*}_k$ are shown in Eq.~\ref{fm:c} and Eq.~\ref{fm:x}.

\subsection{Iterative Contrastive Alignment Framework}
\label{sec:frame}
Considering mini-batch $\mathcal{B}=\{(\textbf{T},\textbf{I})\}$ with $|\mathcal{B}|$ pairs in the whole dataset $\mathcal{D}$, the multimodal abstractive summarization task can be framed as a seq2seq learning problem with extra iterative objectives in this work (more details in Figure~\ref{fig:frame}). The \icaf with a \emph{modified encoder} and a \emph{vanilla transformer decoder} is the backbone architecture, where the model receives the utterances and image as input and outputs a corresponding textual summary $\{{\rm <sos>}, ...,w_i,...,{\rm <eos>}\}$. The encoder is stacked with K-layers RA, interacting cross-modal features (soft-alignment) and passing to the next step.
\begin{equation}
    \begin{aligned}
    \textbf{C}^{T}_k, \textbf{T}_k &= \textbf{RA}(\textbf{T}_{k-1}, \textbf{I}) \\
    \textbf{C}^{I}_k, \textbf{I}_k &= \textbf{RA}(\textbf{I}_{k-1}, \textbf{T})
    \end{aligned}
\end{equation} where $\textbf{C}^{T}_k$ and $\textbf{C}^{I}_k$ indicate step-wise alignment features of text $\textbf{T}$ and image $\textbf{I}$, respectively. $\textbf{T}_k$ and $\textbf{I}_k$ are the $k^{th}$ iterative knowledge in the encoder. In the beginning, $\textbf{T}_0=\textbf{T}$, $\textbf{I}_0=\textbf{I}$ represent the fragments embedding detailed in Section~\ref{sec:represent}. Following the motivation aforementioned, we expect that the corresponding image and text pairs have high consistency, while the irrelevant pairs are hard to capture, especially those fine-grained interplaying. To achieve this goal, we accumulate the contrastive losses advised by infoNCE \cite{oord2018representation} in each RA layer directly.
\begin{equation}
    \begin{aligned}
    \mathcal{L}^{\mathcal{B}}_{T2I} &= -\sum_{i=1}^{|\mathcal{B}|}\sum_{k=1}^{K}{\rm log} \frac{e^{{\rm sim}\left( \hat{f}_{pl}(\textbf{C}_{k,i}^T), \hat{f}_{pl}(\textbf{I}_{i})\right)/\tau}}{\sum^{L}_{j=1} e^{{\rm sim}\left(\hat{f}_{pl}(\textbf{C}_{k,i}^T), \hat{f}_{pl}(\textbf{I}_{j})\right)/\tau}} \\
    \mathcal{L}^{\mathcal{B}}_{I2T} &= -\sum_{i=1}^{|\mathcal{B}|}\sum_{k=1}^{K}{\rm log} \frac{e^{{\rm sim}\left(\hat{f}_{pl}(\textbf{C}_{k,i}^I), \hat{f}_{pl}(\textbf{T}_{i})\right)/\tau}}{\sum^{L}_{j=1} e^{{\rm sim}\left(\hat{f}_{pl}(\textbf{C}_{k,i}^I), \hat{f}_{pl}(\textbf{T}_{j})\right)/\tau}}
    \end{aligned}
\end{equation} where $\tau$ is a temperature hyperparameter and ${\rm sim}()$ is the cosine similarity between two input vectors. Function $\hat{f}_{pl}$ denotes the pooling strategy, i.e., mean/max pooling. The pairs $(\textbf{C}^T_{k,i}, \textbf{I}_i)$ and $(\textbf{C}^I_{k,i}, \textbf{T}_i)$ are positive samples of each other. For $\mathcal{L}^{\mathcal{B}}_{T2I}$ and $\mathcal{L}^{\mathcal{B}}_{I2T}$, these contrastive learners interact within the encoder to emphasize on multimodal pairs while keep text\&image which have similar content but opposite labels away from each other in the embedding spaces. In this way, \icaf is able to optimize the contrastive loss directly by supervising the learning of image-text correspondences at each recurrent alignment layer, which is expected to help the model yield higher-quality alignment in each step.

Following the encoder with multiple RA layers, only the text-target alignment features $\textbf{C}^{T}_K$ is feedforwarded to the decoder, as the Key and Value. Taking the predicted tokens as Query, the standard transformer decoder layer is able to reconstruct the target summary. Compared with the input text, it is shorter but integrates multimodal information better. 

\subsection{Loss Function}
\label{sec:objectives}
The main goal of \icaf is to generate the target summary from input sentences and image, based on seq2seq style framework. Therefore, the reconstruction loss is used to learn the optimal model parameters $\theta$, minimizing the negative log-likelihood:
\begin{equation}
    \mathcal{L}_{gene}^{\mathcal{B}}=-\sum_{\mathcal{B}}\sum_{i}{\rm log} p(w_i|w_{1:i-1}, \textbf{T}, \textbf{I};\theta)
\end{equation} where $w$ denotes the words in the target summary and stops when special token ${\rm <eos>}$ is generated.

\begin{figure}[t]
\centering
\includegraphics[width=\linewidth]{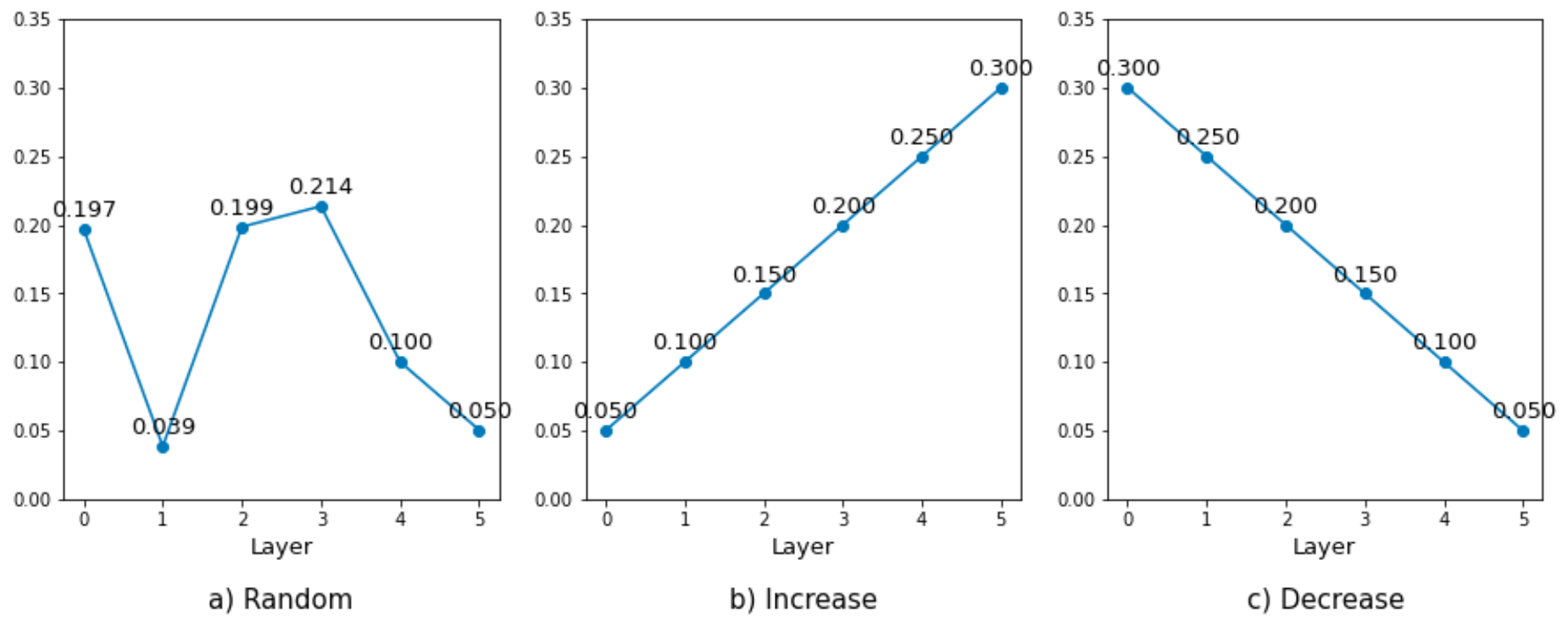}
\caption{\label{fig:beta} Types of static parameters, $\beta_1$ and $\beta_2$.}
\end{figure}

With the auxiliary contrastive losses, the total loss function is defined as Eq.~\ref{fm:loss} and $||\cdot||_2$ denotes the L2 normalization for model parameters $\theta$.
\begin{equation}
    \mathcal{L} = -\frac{1}{|\mathcal{D}|}\sum_{\mathcal{B} \in \mathcal{D}} (\mathcal{L}_{gene}^{\mathcal{B}} + \beta_1\mathcal{L}^{\mathcal{B}}_{I2T} + \beta_2\mathcal{L}^{\mathcal{B}}_{T2I} ) + ||\theta||_2
\label{fm:loss}
\end{equation} where $\beta_1$ and $\beta_2$ are optional static parameters, which include three types (see Figure~\ref{fig:beta}): a) ${\rm Random}$, b) ${\rm Increase}$ and c) ${\rm Decrease}$. Their values range from 0 to 0.3.
% figure 3

% section-3
\section{Experiments}
% table 1
\begin{table*}[ht]
\caption{\label{table:overall} Performance of \icaf and baselines on MSMO Dataset with ROUGE/Relevance/Human Evalution.}
\centering
\small{
        \renewcommand\arraystretch{1.5}
        \begin{tabular}{ccclcccccccc}
        \toprule
        \multirow{2}{*}{\textbf{Type}}   & \multicolumn{2}{c}{\textbf{Resource}} & \multirow{2}{*}{\textbf{Methods}} & \multicolumn{3}{c}{\textbf{ROUGE}}               & \multicolumn{3}{c}{\textbf{Relevance}}           & \multicolumn{2}{c}{\textbf{Human}} \\
                                         & text-only   & image-text   &        & R-1  & R-2 & R-L & Average & Extrema & Greedy & F  & R  \\
        \midrule
        \multirow{5}{*}{\begin{tabular}[c]{@{}c@{}}Pre-Trained \& \\ Fine-tuning \end{tabular}}   & \checkmark  &   & GPT-2 (2019) & 42.16  & 31.16  &40.01 & 0.391 & 0.289 & 0.401 & 3.44 & 3.46   \\
                                                                                 & \checkmark  &   & UniLM (2020) & 47.82  & 34.78  & 44.68 & 0.412 & 0.310 & 0.409 & \textbf{3.52} & 3.39   \\
                                            \cline{2-12}
                                                                                 &    & \checkmark & VLBERT (2020)  & 52.72 & 33.09 & 48.11 & 0.407 & 0.302 & 0.419 & 3.32 & 3.50   \\
                                                                                 &    & \checkmark & UNITER (2020)  & 54.17 & 33.15 & 48.78 & 0.431 & 0.337 & 0.443 & 3.42 & 3.57   \\
                                                                                 &    & \checkmark & OSCAR (2021)  & 52.29 & 32.77 & 47.82 & 0.409 & 0.312 & 0.422 & 3.40 & 3.60   \\
        \midrule
        \midrule
        \multirow{6}{*}{\begin{tabular}[c]{@{}c@{}}Training-\\ from-scrach\end{tabular}} & \checkmark  &  & S2S (2015)   & 29.84   & 11.05   & 26.68  & 0.187 & 0.171 & 0.259 & 2.98 & 3.01\\
                                                                                 & \checkmark  &  & S2S+Attn (2017)  & 32.32 & 12.48 & 29.65 & 0.206 & 0.182  & 0.287 &3.24 & 3.36 \\
                                                                                 & \checkmark  &  & PointerNet (2017) & 34.78 & 13.1 & 32.24 & 0.231 & 0.208 & 0.31 & 3.21 & 3.57  \\
                                           \cline{2-12}
                                                                                 &    & \checkmark & Db-Attn (2017) & 41.78 & 27.81 & 40.28 & 0.362 & 0.269 & 0.389 & 3.13 &3.27 \\
                                                                                 &    & \checkmark & LAMS (2020)  & 47.28  & 30.11 & 45.21 & 0.392 & 0.287 & 0.401 & 3.32 & 3.41 \\
                                                                                 &    & \checkmark & MSGMR (2021)  & 53.22  & 35.48 & 48.23 & 0.443 & \textbf{0.377} & 0.448 & 3.48 & 3.50 \\
                                                                                 &    & \checkmark  & \icaf (ours)  & \textbf{56.11} & \textbf{36.97} & \textbf{49.71} & \textbf{0.452} & 0.369 & \textbf{0.478} & \textbf{3.52}    & \textbf{3.61} \\
        \bottomrule 
        \end{tabular}   
}
\end{table*}

\subsection{Dataset and Metrics Description}
To validate the performance of \icaf on MSMO \cite{zhu-etal-2018-msmo}, this paper compares both \emph{Training-from-scrach} and \emph{Pre-trained then Fine-tuning} baselines, due to the prominent advancement of the pre-train + fine-tune paradigm in both computer vision and natural language processing. This dataset comprises 240,000 image-text pairs for training and 30,000 pairs for both validation and testing from internet news articles with numerous picture-captions. In this paper, the maximum length of utterances is set to 500. Text longer than the maximum length is cutted off. We use an open access tool\footnote{https://github.com/zegami/image-similarity-clustering} to find the most consistent image since the vision input is only one picture in our proposed \icaf.

The evaluation metrics are calculated between the generated summary and the ground-truth by a public tool\footnote{https://github.com/rsennrich/Bleualign}. 
\begin{itemize}[]
\setlength{\itemsep}{3pt}
\setlength{\parsep}{1pt}
\setlength{\parskip}{1pt}
\item \textbf{ROUGE} \cite{lin-2004-rouge}, the standard metric to calculate the scores between the generated summary and reference sentences by using the recall and precision of n-gram overlaps. 
\item \textbf{Relevance} \cite{hashimoto-etal-2016-word}, we use embedding-based metrics to evaluate the summary relevance. In particular, \textbf{Embedding Average}, \textbf{Embedding Extrema} and \textbf{Embedding Greedy} are used to measure the effect of the generated summary.
\item \textbf{Human}, twenty-four native speakers were asked to compare the generated summary with the ground-truth, by rating the fluency (F) and relevance (R) scores, on a scale of 0-4. The male-to-female ratio of 1:1.
\item \textbf{M}$_{\rm sim}$ \cite{sanabria18how2}, an image-text relevance metric calculates the maximum similarity between the image and the generated summary. This metric is only used for ablation analysis for additional contrastive losses.

\end{itemize}

\subsection{Experimental Settings}
We set the word embedding size $D$ to 768 and the limited vocabulary size to 30,004 with four extra special tokens (${\rm <unk>,<pad>,<sos>}$ and ${\rm <eos>}$). We also use dropout with probability equals to 0.3. The number of layers $K$ in \icaf is set to 6. The batch size is up to 128 limited by the GPU (Nvidia 3090 with 24GB VRAM) and the overall parameters are trained for 30 epochs for all the baselines except only 5 epochs for fine-tuning baselines. We use mean pooling, which is verified as the most effective way \cite{Reimers_2019} than Max pooling or [CLS]. We halve the learning rate when development performance worsens. For other hyperparameters, the optimal settings are: similarity scale factor $\gamma=-0.15$, softmax smoothness parameter $\lambda$=6 and temperature parameter in infoNCE $\tau=0.1$. The static parameters $\beta_1$ for $\mathcal{L}^{\mathcal{B}}_{I2T}$ and $\beta_2$ for $\mathcal{L}^{\mathcal{B}}_{T2I}$ in loss function (Eq~\ref{fm:loss}) are both the ${\rm Increase}$, whose values are shown in Figure~\ref{fig:beta}. We further evaluate the performance by using different parameters combination in Section~\ref{sec:ab-anaylsis}. 

\subsection{Quantitative Results}
The performance of this \icaf model and baselines are compared from three perspectives in this work. We demonstrate that the \icaf model is not only superior to de novo training approaches, but also advances pre-trained then fine-tuning paradigms.

\noindent \textbf{Comparisons with Training-from-scratch Methods.}
These models are trained from scratch by randomly initialize the parameters, which can divided into two categories: 1) single-modal (text) baselines and 2) multimodal (text-image) baselines. Category 1 includes traditional methods such as S2S (RNN model) \cite{rush-etal-2015-neural} and S2S+Attn (S2S with attention). PointerNet \cite{DBLP:conf/acl/SeeLM17} learns the conditional probability of the output sequence with elements that are discrete tokens corresponding to the positions in the input sequence. Category 2 is directly related to our proposed \icaf. Doubly-Attn \cite{calixto-etal-2017-doubly} uses various attention to fuse modalities and summarize the input ultimately. MSGMR \cite{zhu2020multimodal} with the guidance of multimodal reference uses the losses from the summary generation and image selection. A location-aware approach for multimodal summarization (LAMS) based on Transformer \cite{DBLP:conf/aaai/ZhangWSY21} is used to illustrate that the performance of \icaf is not resulted from the transformer model. Some of our baselines are advised by the open source website -- NLPedia\footnote{http://explainaboard.nlpedia.ai/}. 

For \emph{Training-from-scratch}, the comparisons between \icaf and baselines are shown in Table \ref{table:overall} (bottom subtable). The results show the \icaf model outperforms all the baselines on ROUGE, Embedding-based metrics and Human evaluation with the only exception of Embedding Extrema. The results also show that adding recurrent alignment module and contrastive learning based auxiliary losses can improve the performance of the generative framework. Through the comparison of the top and bottom parts in this subtable, we can see that using multimodal data as input is better than pure text. Multimodal inputs can provide more references for summary generation. \icaf outperforming LAMS shows that the proposed framework can fuse multimodal references and generate better, not because of the superiority of the transformer model itself.

\noindent \textbf{Comparisons with Pre-train \& Fine-tune Methods.}
For the sake of fairness, this paper updates 5 epochs for the textual pre-trained models GPT-2\footnote{Gpt-2 source code from https://github.com/openai/gpt-2.}, UniLM\footnote{UniLM source code from https://github.com/microsoft/unilm} and multimodal pre-trained models VLBERT\footnote{VLBERT pre-trained params from https://github.com/jackroos/VL-BERT}, UNITER\footnote{UNITER params from https://github.com/ChenRocks/UNITER}. The former two have the ability of textual generation by designing model structures (GPT-2) or masked matrix (UniLM), while the latter two have the ability of cross-modal understanding by additional tasks such as masked classification (VLBERT) or matching (UNITER). OSCAR \cite{li2020oscar} is a more recent multimodal pre-training model that employed object tags to match with word tokens and salient regions. The same decoder as \icaf in OSCAR is trained for an extra 25 epochs.

The comparisons between \icaf and baselines of \emph{Pre-Trained+Fine-tuning} are shown in the top of Table \ref{table:overall}. It can be illustrated that no matter whether compared with the single-stream or the multi-streams pre-trained models, the \icaf model always performs better in both metrics and human evaluation. This greatly proves the excellence of \icaf. Meanwhile, the baseline of text-only pre-trained model is better in summary fluency, while the multimodal pre-trained model is excellent in correlation. 

\subsection{Ablation Analysis}
\label{sec:ab-anaylsis}
\begin{table}[t]
\caption{\label{table:hyper} The effect of the hyper parameters, which are $\gamma, \tau$, $K$ and types of $\beta$.}
\centering
\small{
\renewcommand\arraystretch{1.4}
\begin{tabular}{llll}
\toprule
Methods                & ROUGE & Relevance & M$_{sim}$\\ \hline
\icaf                  & 47.60 & 0.433     & 0.428 \\ \hline \hline
\multicolumn{4}{l}{Default $\gamma=-0.15$}         \\ \hline
\icaf ($\gamma=-0.20$) & 47.31 & 0.427     & 0.433 \\
\icaf ($\gamma=-0.10$) & 47.51 & 0.430     & 0.391 \\
\icaf ($\gamma=-0.05$) & 46.18 & 0.418     & 0.347 \\
\icaf ($\gamma=0$)     & 45.72 & 0.414     & 0.302 \\ \hline \hline
\multicolumn{4}{l}{Default $\lambda=6$}            \\ \hline
\icaf ($\lambda=3$)    & 42.16 & 0.382     & -     \\
\icaf ($\lambda=9$)    & 44.38 & 0.393     & -     \\ \hline \hline
\multicolumn{4}{l}{Default $\tau=0.1$}          \\ \hline
\icaf ($\tau=0.01$)    & 46.18 & 0.411     & -     \\
\icaf ($\tau=0.2$)     & 40.66 & 0.378     & -     \\ \hline \hline
\multicolumn{4}{l}{Default K=6 }                   \\ \hline 
\icaf (K=3)            & 45.57 & 0.409    & -     \\
\icaf (K=9)            & 47.64 & 0.432    & -     \\ \hline \hline
\multicolumn{4}{l}{Default $\beta_1 \Rightarrow {\rm Increase} \quad \& \quad \beta_2 \Rightarrow {\rm Increase}$ } \\ \hline 
\icaf ($\beta_1 \Rightarrow {\rm Random}$)            & 46.44 & 0.421    & 0.419     \\
\icaf ($\beta_2 \Rightarrow {\rm Random}$)            & 44.71 & 0.396    & 0.331     \\
\icaf ($\beta_1\Rightarrow {\rm Random}$ & \multirow{2}{*}{43.63} & \multirow{2}{*}{0.389} & \multirow{2}{*}{0.328} \\
$\quad\qquad \beta_2\Rightarrow {\rm Random}$) & & & \\ \hline
\icaf ($\beta_1 \Rightarrow {\rm Decrease}$)            & 46.07 & 0.414    & 0.397     \\
\icaf ($\beta_2 \Rightarrow {\rm Decrease}$)            & 41.22 & 0.374    & 0.316     \\
\icaf ($\beta_1\Rightarrow {\rm Decrease}$ & \multirow{2}{*}{40.12} & \multirow{2}{*}{0.362} & \multirow{2}{*}{0.305} \\
$\quad\qquad \beta_2\Rightarrow {\rm Decrease}$) & & & \\

\bottomrule
\end{tabular}}
\end{table}

In this section, we first evaluate the rationality of the default hyperparameters settings and the chosen types of static factors in loss function, and then verify the effectiveness of the proposed recurrent alignment layer and cross-modal contrastive learning.

\textbf{Effect of similarity scale factor}, $\gamma$. In the recurrent alignment layer, the closer the cosine similarity is  to 1 , the higher the semantic correlation. Setting the default $\gamma$ as a negative value aims to filter more semantical information. As shown in Table~\ref{table:hyper}, a larger or smaller $\gamma$ value decreases the performance of \icaf. We can also see that when this threshold increases, the similarity between the selected image and target summary will decrease (the value of M$_{sim}$ becomes smaller). Experiments show that this is an extremely sensitive parameter for different situation or datasets.

\textbf{Effect of temperature parameter}, $\tau$. It is a hyperparameter for contrastive loss. Similar to \cite{gao2021simcse}, we also find that $\tau=0.1$ is the most effective value for Mean pooling strategy.

\textbf{Effect of layers}, $K$. For \icaf, we gradually increase K from 3 to 9 and evaluate them on the benchmark datasets. Due to space limitation, we only report the results on 3 and 9 in Table~\ref{table:hyper}. We can observe that for all variants, K=9 can achieve better but limited performance than K=3 or K=6. The model improvement in exchange for more model parameters is minimal in this analysis. 

\textbf{Effect of choosen types of $\beta_1$ and $\beta_2$.} Because the cross-modal contrastive loss is accumulated with the iterations of RA layers, i.e., the loss of $k^{th}$ layer will be back propagated to the parameters of 1 to k layers. Therefore, using the $\beta$ of ${\rm Increase}$ type is conducive to the uniform optimization of model parameters intuitively. The experimental results also show that the performance degrades to a certain extent by replacing $\beta_1$ and $\beta_2$ with ${\rm Random}$ and ${\rm Decrease}$.

\begin{table}[]
\caption{\label{table:icaf} The effect of iterative contrastive alignment process.}
\centering
\small{
\renewcommand\arraystretch{1.4}
\begin{tabular}{llll}
\toprule
Methods                & ROUGE & Relevance & M$_{sim}$\\ \hline
\icaf                  & 47.60 & 0.433     & 0.428 \\ 
MSGMR                  & 45.64 & 0.423     & 0.264 \\ \hline
-$\mathcal{L}^{\mathcal{B}}_{T2I}$ & 39.27 & 0.352     & 0.302 \\
-$\mathcal{L}^{\mathcal{B}}_{I2T}$& 45.48 & 0.411     & 0.387 \\
-$\mathcal{L}^{\mathcal{B}}_{T2I} \& \mathcal{L}^{\mathcal{B}}_{I2T}$ & 36.71 & 0.341     & 0.237 \\ \bottomrule
\end{tabular}}
\end{table}
% Table 4
\begin{table*}
\caption{\label{table:case} A case study. The references of original utterances and images are given and the corresponding summary generated by both \icaf and baselines are given as well.}
\renewcommand{\arraystretch}{1.4}
\begin{tabular}{cl}
\toprule
\textbf{\begin{tabular}[c]{@{}c@{}}Original Textual\\ Reference\end{tabular}}          & \begin{tabular}[c]{@{}l@{}}\qquad Li Na is destined for more grand slams and a place in the Hall of Fame after the Australian Open win cemented her \\ position as the most influential female player of the last decade.  Li's final victory over Slovak Dominika Cibulkova gave \\ her a second major following her 2011 French Open title, a triumph that made her Asia's first singles grand slam champion.\\ \qquad Li was worthy of a place alongside her in the Hall of Fame. It's not only about winning grand slams its about the \\ influence that you have in tennis. Stacey said she is the most influential women's tennis player in the last 10 years, with \\ what she has done for global tennis so absolutely 100 percent.\\ \qquad She is right up there with them too. There was always a little gap before you said Li Na's name but now I think she is \\ right up there with all of them after the type of tennis she played at the Australian Open.\end{tabular} \\ 
\hline
\multirow{6}{*}{\textbf{\begin{tabular}[c]{c}Original Image\\ Reference\end{tabular}}} & \begin{minipage}[b]{0.2\columnwidth}
		\centering
		\raisebox{-1\height}{\includegraphics[width=8\linewidth]{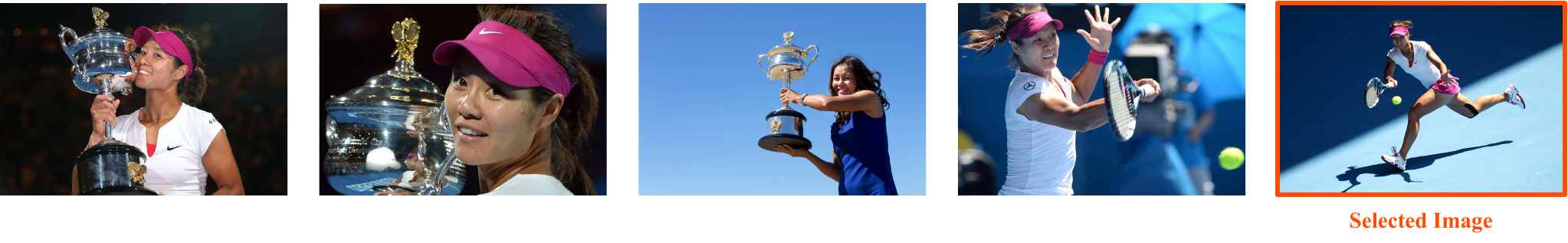}}
	\end{minipage} \\
\hline
\multirow{10}{*}{\textbf{\begin{tabular}[c]{@{}c@{}}Generated \\ Summary\end{tabular}}} & \textbf{PointerNet}:   Li was worthy of  Australian Open and the most influential women's tennis player. \\ \cline{2-2} 
& \begin{tabular}[c]{@{}l@{}}\textbf{LAMS}:  After \textcolor{blue}{win} Slovak Dominika Cibulkova, Li Na has been the most influential \textcolor{blue}{women's player} of the last decade after \\ 2011 French Open \textcolor{blue}{title}.\end{tabular}\\ \cline{2-2} 
& \begin{tabular}[c]{@{}l@{}}\textbf{UniLM}: Li was worthy of the Hall of Fame. It's about the position of \textcolor{blue}{the} influential player and win the Australian Open \\ game.\end{tabular} \\ \cline{2-2} 
& \begin{tabular}[c]{@{}l@{}}\textbf{UNITER}:  Li Na is playing tennis with Slovak Dominika Cibulkova. After wining the Australian Open, she has \textcolor{blue}{be} the 10 \\ years most influential women's tennis player.\end{tabular} \\ \cline{2-2} 
& \begin{tabular}[c]{@{}l@{}}\textbf{\icaf}: \textcolor{red}{Li Na} has been \textcolor{red}{the most influential women's tennis player} following her \textcolor{red}{2011 French Open} and \textcolor{red}{Australian Open win}. \\ It's not only about  the \textcolor{red}{first Asia's singles grand slam champion} its about \textcolor{red}{the influence} that she has \textcolor{red}{in tennis}.\end{tabular}    \\ 
\bottomrule
\end{tabular}
\end{table*}

\textbf{Effect of the recurrent alignment layer.} The aggregation function $f()$ in Eq~\ref{fm:c} and the cross-modal contrastive losses in Eq~\ref{fm:loss} are essential for the proposed iterative contrastive alignment process. We enumerate three derivations and compare them with \icaf in Table~\ref{table:icaf}: 1) removing $\mathcal{L}^{\mathcal{B}}_{T2I}$ directly affects the alignment of image alignment to text; 2) removing $\mathcal{L}^{\mathcal{B}}_{I2T}$ has little impact, the generation process can still be carried out, and the image-text matching is still better than MSGMR; 3) removing both $\mathcal{L}^{\mathcal{B}}_{T2I}$ and $\mathcal{L}^{\mathcal{B}}_{I2T}$ directly degenerates the model into a basic transformer model, except for the encoder layer. We can observe that RA achieves substantially better performance than any other derivations.

\subsection{Case Study}
\label{sec:case}
We demonstrate a successful example of the MSMO dataset in this section. Table~\ref{table:case}, which comprises the original textual paragraphs, photos, and the target generated summary containing the information (both baselines and \icaf). The summary gets textual references (red words) from the textual reference, as can be observed intuitively. The output provided by \icaf is longer (contains more information) and free of evident errors when compared to the textual summary generated by the baselines (the blue words indicates possible errors in the results).

% section-4
\section{conclusion}
We introduce an end-to-end summarization generation framework by using multimodal information instead of pure text features. This work addresses the multimodal abstractive summarization problem by adding two auxiliary contrastive goals and reconstructing cross-modal alignment layer, aiming to summarize multimodal data and measure similarity simultaneously. The proposed \icaf model shows that the RA layers can extract knowledge from multimodal data and complete the abstractive summarization by using recurrent contrastive learning. We also demonstrate that our model outperforms benchmark systems in ROUGE, relevance scores and human evaluation, creating a new state-of-the-art.

\section*{Acknowledgment}
This work is supported by Ningbo Science and Technology Bureau under Service Industry S\&T Programme with project code 2019F1028, Major Projects Fund with project code 2021Z089, and Ministry of Science and Technology of China under Grant 2020AAA0104200.

\clearpage
\bibliographystyle{IEEE}
\bibliography{references}

\vspace{12pt}

\end{CJK}
\end{document}